\documentclass[10pt,twocolumn,letterpaper]{article}

\usepackage{cvpr}              
\usepackage{graphicx}

\usepackage{amsfonts}
\usepackage{amsthm}
\usepackage{amsmath}
\usepackage{amssymb}
\usepackage{bm}
\usepackage{xspace}
\usepackage{nicefrac}
\usepackage{cases}
\usepackage{mathtools}
\usepackage{booktabs}
\usepackage{threeparttable}
\usepackage{enumitem}
\usepackage{colortbl}
\usepackage{multirow}
\usepackage{adjustbox}

\usepackage{microtype}
\usepackage[normalem]{ulem}
\usepackage{xcolor}
\usepackage{color}

\usepackage{pifont}
\usepackage{subcaption}
\usepackage{draftfigure}
\usepackage{lipsum}

\theoremstyle{plain}

\theoremstyle{definition}

\theoremstyle{remark}

\usepackage{algorithm}
\usepackage{algorithmic}

\newcommand{\cmark}{\ding{51}}%
\newcommand{\xmark}{\ding{55}}%

\newcommand{\method}{MetaCompress\xspace}
\newcommand{\cls}{[\texttt{CLS}]\xspace}

\definecolor{cvprblue}{rgb}{0.21,0.49,0.74}
\usepackage[pagebackref,breaklinks,colorlinks,allcolors=cvprblue]{hyperref}

\usepackage[capitalize,noabbrev]{cleveref}

\title{Rethinking Token Reduction for Large Vision-Language Models}

\author{
    \textbf{Yi Wang$^{1\ast}$,
    Haofei Zhang$^{2,3\ast}$,
    Qihan Huang$^{1}$,
    Anda Cao$^{1}$,
    Gongfan Fang$^{5}$,} \\
    \textbf{Wei Wang$^{6}$,
    Xuan Jin$^{6}$,
    Jie Song$^{4}$,
    Mingli Song$^{1,2,3,4}$,
    Xinchao Wang$^{5\dagger}$} \\
    \vspace{-0.8em}\\
    $^{1}$College of Computer Science and Technology, Zhejiang University \\
    $^{2}$State Key Laboratory of Blockchain and Data Security, Zhejiang University \\
    $^{3}$Hangzhou High-Tech Zone (Binjiang) Institute of Blockchain and Data Security \\
    $^{4}$School of Software Technology, Zhejiang University \\
    $^{5}$National University of Singapore ~\quad $^{6}$Alibaba Group \\%
    \vspace{-2.5em}
}

\begin{document}
    \maketitle
    
    \makeatletter
    \renewcommand{\thefootnote}{\fnsymbol{footnote}}
    \setcounter{footnote}{0}
    \footnotetext[1]{Equal Contribution. Email: y\_w@zju.edu.cn, haofeizhang@zju.edu.cn}
    \footnotetext[2]{Corresponding Author. Email: xinchao@nus.edu.sg}
    \renewcommand{\thefootnote}{\arabic{footnote}}
    \makeatother
    
    \begin{abstract}

\vspace{-0.5em}
Large Vision-Language Models (LVLMs) excel in visual understanding and reasoning, but the excessive visual tokens lead to high inference costs. Although recent token reduction methods mitigate this issue, they mainly target single-turn Visual Question Answering (VQA), leaving the more practical multi-turn VQA (MT-VQA) scenario largely unexplored. MT-VQA introduces additional challenges, as subsequent questions are unknown beforehand and may refer to arbitrary image regions, making existing reduction strategies ineffective. Specifically, current approaches fall into two categories: prompt-dependent methods, which bias toward the initial text prompt and discard information useful for subsequent turns; prompt-agnostic ones, which, though technically applicable to multi-turn settings, rely on heuristic reduction metrics such as attention scores, leading to suboptimal performance. In this paper, we propose a learning-based prompt-agnostic method, termed \method, overcoming the limitations of heuristic designs. We begin by formulating token reduction as a learnable compression mapping, unifying existing formats such as pruning and merging into a single learning objective. Upon this formulation, we introduce a data-efficient training paradigm capable of learning optimal compression mappings with limited computational costs. Extensive experiments on MT-VQA benchmarks and across multiple LVLM architectures demonstrate that \method achieves superior efficiency–accuracy trade-offs while maintaining strong generalization across dialogue turns. Our code is available at \url{https://github.com/MArSha1147/MetaCompress}.

\end{abstract}

    \vspace{-1.3em}
    \section{Introduction} \label{sec:intro}
\vspace{-0.2em}

LVLMs~\cite{liu2023llava,liu2023improvedllava,liu2024llavanext,internlmxcomposer,internlmxcomposer2_5,achiam2023gpt,bai2025qwen2} have become powerful AI systems enabling natural human interaction with visual data such as images and videos. They encode both textual and visual modalities into tokens jointly processed by a unified Large Language Model~(LLM). Recent works~\cite{liu2024llavanext,internlmxcomposer2_5} further extend LVLMs toward multi-scale visual inputs that integrate both global and local tokens to enhance visual understanding. However, visual tokens greatly increase computation and memory costs, as token numbers grow by thousands and attention scales quadratically with sequence length~\cite{dao2022flashattention,katharopoulos2020transformers}, making low-latency or resource-constrained deployment challenging~\cite{yao2024minicpm}.

Although numerous token reduction techniques~\cite{chen2024image,ye2024fitprune,shang2024llava} have been proposed and have achieved considerable success, they are primarily developed for single-turn VQA.
Meanwhile, the more realistic MT-VQA setting, which involves multi-round conversational question answering, remains largely underexplored thus far.
Compared to single-turn VQA, which focuses on answering one-shot questions and can greedily discard image tokens irrelevant to the current query, MT-VQA poses additional challenges due to its open-ended nature.
In MT-VQA, future questions are entirely unpredictable, and their relevant regions may arise anywhere in the image, making existing token reduction methods difficult to apply directly.

Existing token reduction methods can be broadly categorized into prompt-dependent and prompt-agnostic approaches. Prompt-dependent methods, such as FastV~\cite{chen2024image}, retain tokens that are highly relevant to only the first-turn question prompt. This strategy may inadvertently discard visual information that could be crucial for answering subsequent questions; for example, the first question might focus on the foreground while later questions may reference the background. In contrast, prompt-agnostic approaches like PruMerge~\cite{shang2024llava} reduce tokens solely based on attention scores within the image sequence itself, which are technically applicable to multi-turn interactions. However, a critical limitation of existing prompt-agnostic methods lies in their reliance on heuristic reduction criteria derived from human priors, and the lack of theoretical guidance, often resulting in suboptimal performance.

In response to this, we propose a learning-based prompt-agnostic token reduction approach, termed \method, which overcomes the drawbacks of heuristic designs. To achieve this, a key question is how the learning objective should be defined. By analyzing the reduction formats of current approaches, including both pruning and merging, we find that they can be unified by formulating the visual token reduction task as an optimization problem. The goal is to identify an optimal compression mapping of the input visual tokens, under conditions such as language conditioning in prompt-dependent approaches and image-only conditioning in prompt-agnostic approaches, so that the model’s responses exhibit minimal discrepancy after token reduction.

Based on this formulation, we first simplify the problem by learning an optimal compression matrix for each image and conduct a preliminary investigation into the guiding role of attention information, as commonly employed in previous methods. Surprisingly, our findings reveal that the tokens retained by the learned matrix do not exhibit an obvious relationship with the heuristic attention cues commonly used in prior methods, such as \cls token attention and prompt-token attention, further validating that heuristic reduction criteria are suboptimal.

Furthermore, to fully implement \method, a practical challenge arises from the need to generate multiple compression matrices, since actual image inputs can vary in resolution.
And learning specific compression matrices for every possible resolution is not an especially elegant or practical solution.
To address this, we ultimately design to learn a compression matrix generator compatible with dynamic resolutions, trained in a data-efficient paradigm.
Extensive experiments on three MT-VQA benchmarks using five LVLM architectures demonstrate that \method outperforms existing token reduction methods while achieving high computational efficiency.

The contributions of our paper are summarized as follows:
\begin{itemize}
    \item We first explore token reduction in the MT-VQA scenario, revealing that heuristic methods relying on visual token attention scores are suboptimal.
    \item We propose \method, a novel learning-based and prompt-agnostic token reduction method, overcoming the reliance on suboptimal heuristic reduction criteria.
    \item \method leverages a data-efficient training paradigm to learn the optimal compression mapping for visual sequences, demonstrating the effectiveness and efficiency through extensive experiments.
\end{itemize}
    \section{Related Work} \label{sec:related_work}

\subsection{Efficient Large Vision-Language Models} \label{sec:related_work-lvlm}

\textbf{LVLMs.} Transformers~\cite{vaswani2017attention} have unified architectures across language~\cite{devlin2019bert,raffel2020exploring,brown2020language} and vision~\cite{dosovitskiy2021an,touvron2021training,carion2020end,han2025hi,HONG2021103224}, then CLIP~\cite{radford2021learning} bridges both modalities through contrastive pre-training, enabling zero-shot visual understanding. Based on this, LVLMs~\cite{alayrac2022flamingo,yu2022coca,wang2022git,liu2023llava,liu2023improvedllava,liu2024llavanext,internlmxcomposer2_5} integrate visual encoders with large language models to perform multimodal tasks such as captioning and VQA. LLaVA~\cite{liu2023llava,liu2023improvedllava} achieves image-to-text generation by feeding CLIP-encoded visual tokens and language tokens into an LLM~\eg, Llama~\cite{touvron2023llama,touvron2023llama2}, but its fixed global resolution restricts fine-grained perception. Recent models such as LLaVA-NeXT~\cite{liu2024llavanext} and InternLM-XComposer-2.5~\cite{internlmxcomposer2_5} enhance visual understanding by incorporating multi-scale visual inputs that combine global and local tokens, but this substantially increases token counts, leading to heavy memory and computation overhead in multi-head attention and auto-regressive decoding, particularly on resource-constrained devices.

\noindent\textbf{Model Quantization.} To deploy LVLMs to low-memory devices such as mobile while preserving the model's performance, a natural approach is to quantize the model and inference process into 4/8-bit~\cite{dettmers2022gpt3,egashira2024exploiting,yuan2024pbllm} or even 1-bit~\cite{ma2024era}. Another line of work focuses on reducing the computational burden of MHA by employing efficient attention mechanisms~\cite{dao2022flashattention,dao2023flashattention2,xFormers2022,chen2025dparallel} or sparse attention~\cite{child2019generating,xiao2024efficient,wang2026sparsed}. However, quantization methods are limited by optional fine-tuning and hardware support, and more importantly, they do not solve the overall computational inefficiency caused by the increasing number of visual tokens.

\noindent\textbf{Model Pruning.} Model pruning~\cite{Liu2019MetaPruningML,Zhu2021VisualTP,ma2023llmpruner,fang2023depgraph} and knowledge distillation~\cite{Hinton2015DistillingTK,sanh2019distilbert,Chuanpeng2024SurveyKD} methods compress the given model to arbitrary size by removing redundant parameters or transferring knowledge from a large model to a smaller one. These approaches are generally effective at reducing model size and inference cost, but often require careful hyperparameter tuning and expensive retraining procedure.

\subsection{Visual Token Reduction} \label{sec:related_work-token_reduction}
\vspace{-0.2em}
Recent studies show that image representations contain substantial redundancy~\cite{he2022masked,deng2022discovering}, enabling feature reduction without significant performance loss~\cite{bolya2023token,cao2024madtp,Chen2021DistributionKE}.
This observation has motivated the development of token reduction techniques for LVLMs, which can be broadly categorized into prompt-dependent and prompt-agnostic approaches. Prompt-dependent methods, such as FastV~\cite{chen2024image}, identify redundant tokens by measuring their attention to language prompts and remove them at specific layers. FitPrune~\cite{ye2024fitprune} identifies redundant tokens by minimizing the divergence of attention distributions before and after pruning. IVTP~\cite{kai2025ivtp} and TRIM~\cite{song2024less} employ CLIP’s text encoder to guide token reduction. However, these methods are less applicable to general scenarios such as MT-VQA, as they require re-compression for each question. In contrast, prompt-agnostic methods rely solely on image sequences and are technically applicable to MT-VQA tasks. Nevertheless, existing approaches like LLaVA-PruMerge~\cite{shang2024llava} overlook compatibility with modern LVLMs that incorporate multi-scale vision towers (\eg, LLaVA-NeXT). More importantly, these methods rely heavily on heuristic reduction criteria derived from human priors, which often lead to suboptimal performance. To address this, this paper introduces \emph{a novel learning-based, prompt-agnostic token reduction method that avoids heuristic designs} (\eg, attention to \cls or other tokens as reduction guidance, which will be shown suboptimal later in this paper) and can integrate seamlessly with modern LVLMs.

    \section{Preliminaries} \label{sec:method-preliminaries}
In this section, we first give a brief review of the inference process of LVLMs, particularly in the context of multi-turn dialogue scenarios. We then introduce the problem definition of visual sequence compression mapping.

\subsection{Large Vision-Language Models} \label{sec:method-preliminaries-lvlms}
Given an input image $I_\text{IMG}$, LVLMs are required to generate a series of responses $(R_1, \ldots, R_t)$ to user's prompts $(P_1, \ldots, P_t)$.
The image and the language context are tokenized separately by a vision tower $T_\text{IMG}(\cdot)$, \eg, vision Transformer~(ViT)~\cite{dosovitskiy2021an,radford2021learning} and a language tokenizer $T_\text{TXT}(\cdot)$, \eg, SentencePiece~\cite{kudo2018sentencepiece}.
The tokenized image and language sequence are then embedded into a common space by a vision projector $V_\text{IMG}(\cdot)$ and an embedding layer $E_\text{TXT}(\cdot)$, producing $X_\text{IMG}$ and $X_\text{TXT}$, respectively.

To fully capture detailed information, current prevalent LVLMs, such as LLaVA-NeXT and InternLM-XComposer-2.5, employ a ViT to encode images from both global and local views, generating multi-scale visual sequences.
Despite enhancing the model's capability to capture the details of the image, such an approach significant increase the token number, severely impairing the inference efficiency due to the $O(n^2)$ computational and memory complexity of MHA~\cite{vaswani2017attention}.

To increase LLMs' inference efficiency, KV cache methods~\cite{liu2024minicache,yang2024pyramidinfer} are proposed for reusing intermediate attentions in the auto-regressive decoder.
Specifically, for generating the $i$-th response token with query $q_i$, the original computation is to concatenate the previous queries for MHA layer
\begin{equation}
    \mathrm{MHA}(Q_i, K_i, V_i) = \sigma\left(\frac{Q_i K_i^\top}{\sqrt{d_k}}\right)V_i \text{,}
\end{equation}
where $\{Q,K,V\}_i = (\{Q,K,V\}_{i-1} | \{q,k,v\}_i)$ are the concatenated inputs, and $\sigma$ denotes the row-wise Softmax operation.
To decrease the computation complexity, KV caches store the intermediate key-value pairs $\{K,V\}_{i-1}$ and only compute the attention for $q_i$:
\begin{equation}
    \mathrm{MHA}(q_i, k_i, v_i) \!=\! \sigma\left(\!\frac{q_i (K_{i-1}|k_n)^\top}{\sqrt{d_k}}\!\right)\!(V_{i-1}|v_i) \text{,}
\label{eq:mha}
\end{equation}
which significantly reduce the computation burden.
Such techniques can be seamlessly integrated with the MT-VQA setting, where the caches are reused across multiple turns.

\subsection{Visual Token Reduction} \label{sec:method-preliminaries-comp}
However, the aforementioned cache mechanism is still insufficient to address the memory and computation overhead caused by the large number of image tokens, resulting in an $O(n^2)$ cost for generating the first token and an $O(nT)$ cost for producing $T$ tokens during multi-turn dialogues.

To alleviate the issue, token reduction methods are proposed to compress the image sequence.
For simplicity, we only consider reducing image tokens right before feeding into the LLM, \eg, LLama~\cite{touvron2023llama}, which can be formulated as
\begin{equation}
    \tilde{X}_\text{IMG} = \mathcal{P}_\text{reduce}(X_\text{IMG}|I_\text{IMG}, I_\text{TXT}) \text{,}
\label{eq:mha-cache}
\end{equation}
where guiding information is extracted from the input image $I_\text{IMG}$ and the language context $I_\text{TXT}$.
Depending on whether to rely on the prompt $I_\text{TXT}$, token reduction methods can be categorized into prompt-dependent and prompt-agnostic methods.
However, in real-world applications, LVLMs are often required to respond to multiple prompts.
Prompt-dependent methods, however, tend to bias toward the initial query and discard information beneficial for subsequent turns, leading to suboptimal performance in multi-turn dialogue scenarios. Furthermore, many existing methods requires the intermediate attention matrices in MHA layers to guide token reduction, whereas modern LVLMs commonly employ FlashAttention~\cite{dao2022flashattention,dao2023flashattention2} or Memory-Efficient Attention~\cite{xFormers2022}, which do not support returning them.

To further investigate the optimal token reduction strategy, we first unify the token pruning and merging methods by formulating the reduction process as a linear projection to the input $X_\text{IMG}$:
\begin{equation}
    \tilde{X}_\text{IMG} = PX_\text{IMG}\text{,}
\label{eq:comp_proj}
\end{equation}
where $P\in\mathbb{R}_+^{m\times n}$ ($m \ll n$) is the sparse compression matrix.
In~\Cref{sec:method-optimal_mapping}, we set $P$ as a learnable matrix for each image.
By optimizing $P$ in a data-driven manner, we first explore the relationship between the retained tokens and the attention weights employed by heuristics-designed methods.
Then in~\Cref{sec:method-seq_comp_map}, we present a novel token reduction method that does not rely on the intermediate attention matrices, while can be seamlessly integrated with modern LVLMs.

\section{Which Tokens to Keep?} \label{sec:method-optimal_mapping}

\begin{figure*}[!t]
\centering%
\includegraphics[width=\linewidth]{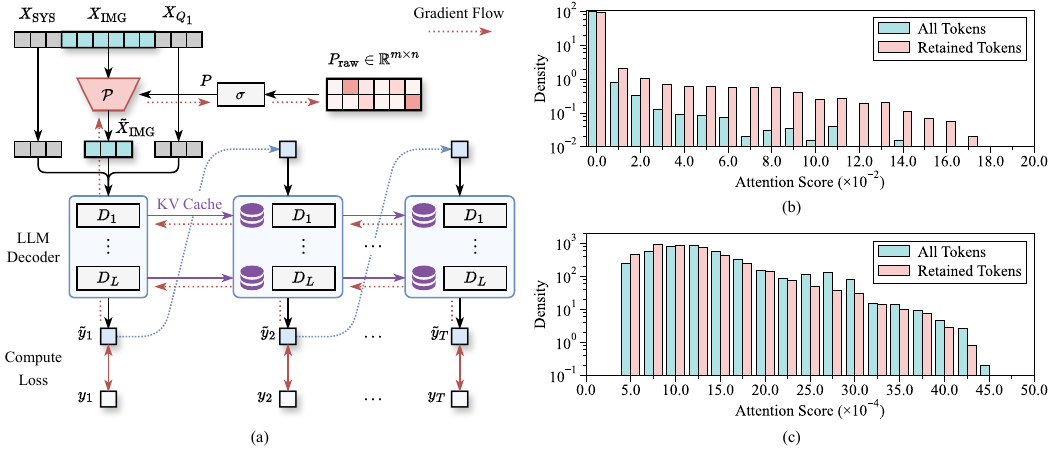}%
\caption{%
    (a) Overall pipeline of the compression projection training process.
    (b) Attention distribution over the \cls token for retained and all visual tokens. The image tokens are extracted from the last layer of the vision tower of LLaVA-1.5-13b running on VQA-v2 dataset.
    (c) Attention distribution over the prompt tokens for retained and all visual tokens. The attention scores are averaged to prompt tokens extracted from the first layer of the LLM decoder.%
}%
\label{fig:comp_one_img}%
\vspace{-1.0em}
\end{figure*}%

To objectively analyze the optimal token reduction scheme without relying on hand-crafted designs, we start by looking at a simpler case: \emph{Given an input image \protect$I_\text{IMG}$ and a conversation context $I_\text{TXT}$, find the optimal compression matrix \protect$P^*$ as defined in~\mbox{\Cref{eq:comp_proj}} so that the response discrepancy between the LLM using the compressed and uncompressed visual sequence is minimized.}

To achieves this, let $P_\text{raw} \in \mathbb{R}^{m\times n}$ be the trainable reduction parameters, with each element independently drawn from Gaussian distribution $\mathcal{N}(0, \sigma_\text{raw}^2)$.
We normalize $P_\text{raw}$ with row-wise Softmax to obtain the compression matrix $P=\sigma(P_\text{raw})$.
Let $p(y|X_\text{IMG}, 
X_\text{TXT})$ denotes the original prediction distribution $T$ tokens 
$y = (y_1, \dots, y_T)$.
Then we force the LLM to generate $T$ tokens 
$p(\tilde{y}|\tilde{X}_\text{IMG}, X_\text{TXT})$ 
using the compressed visual sequence $\tilde{X}_\text{IMG}$.
\Cref{fig:comp_one_img}a illustrates the overall training pipeline, where $P_\text{raw}$ is trained to minimize the KL divergence between the two response distributions 
$\mathcal{L}_\text{pred} = D_\text{KL}(p(y) \| p(\tilde{y}))$ 
and the distribution entropy $\mathcal{L}_\text{entropy} = \frac{1}{m}\sum_{i=0}^m \mathcal{H}(P_{i, :})$.

The training objective is formulated as
\begin{equation}
    P^* = \mathop{\arg\min}_{P_\text{raw}}\mathcal{L}_\text{pred} + \alpha \mathcal{L}_\text{entropy} \text{.}
\label{eq:loss_single_img}
\end{equation}
The training algorithm and detailed implementation are provided in~\Cref{app:details}.

\Cref{fig:comp_one_img}b and \ref{fig:comp_one_img}c visualize the attention distribution over the \cls and prompt tokens, respectively.
Despite that some tokens with high attention to the \cls token are retained (accounting for approximately 1.71\% of the total retained tokens), \emph{the vast majority of the retained tokens are unrelated to their attention scores}, especially with regards to the attention to the language prompts.
More results are in~\Cref{sec:exp-vis}, drawing to the same conclusion.
This observation suggests that using attention scores as guidance for token reduction is suboptimal in the MT-VQA scenario, which explains the experimental results in~\Cref{sec:exp-main}, where token pruning methods such as FastV perform worse than uniform or even random pruning.
Therefore, it is essential to explore a novel token reduction approach that does not rely on heuristic metrics such as attention scores, while being seamlessly compatible with modern LVLMs.

\section{Method} \label{sec:method-seq_comp_map}
Results from~\Cref{sec:method-optimal_mapping} inspire us to construct the compression matrix in a data-driven manner.
To this end, we propose \method, a lightweight module learning the compression matrix $P$ conditioned only on the input image for MT-VQA scenarios.
\Cref{sec:method} details the \method module, \Cref{sec:method-theory} provides theoretical analysis, and \Cref{sec:method-training} presents the optimization objective and training algorithm.

\subsection{\method} \label{sec:method}
Our goal is to learn a compression matrix generator $\mathcal{P}_\text{meta}$ in a data-driven manner, so that the overall prediction discrepancy on the given dataset $\mathcal{D}=\{(I_\text{IMG}^{(i)}, I_\text{TXT}^{(i)})\}_{i=1}^N$ is minimized.
To this end, we propose a lightweight meta generator $\mathcal{P}_\text{meta}$ which computes a compression matrix $P=\mathcal{P}_\text{meta}(X_\text{IMG})$ for each input image $I_\text{IMG}$, independent of the prompt.
One major challenge is that $\mathcal{P}_\text{meta}$ is required to generate the compression matrix $P$, whose shape can adapt to the varying length of $X_\text{IMG}$, thereby accommodating multiple resolution scales for LVLMs such as LLaVA-NeXT and InternLM-XComposer-2.5.

\begin{figure}[!t]
\centering%
\includegraphics[width=\linewidth]{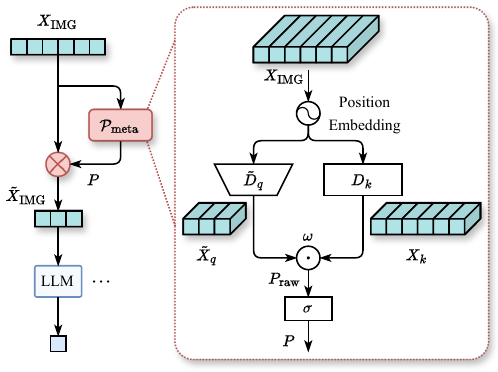}%
\caption{Illustration of our proposed \method, where module $\mathcal{P}_\text{meta}$ generate the compression projection $P$ solely according to the image sequence $X_\text{IMG}$.}%
\label{fig:method}%
\vspace{-1.0em}
\end{figure}

\Cref{fig:method} shows the overall architecture of $\mathcal{P}_\text{meta}$, which consists of a position embedding layer, a query down-sample projection $\tilde{D}_q$, a key projection $D_k$, and a weighted inner product layer.
The core idea is to compute the inner product between the spatially down-sampled queries $\tilde{X}_q\in\mathbb{R}^{m\times d_c}$ and keys $X_k \in \mathbb{R}^{n\times d_c}$ to get the compression matrix $P\in\mathbb{R}^{m\times n}$.
Here, queries
\begin{equation}
    \tilde{X}_q = \tilde{D}_q(X_\text{IMG} + E_\text{pos}) = \mathrm{Pool}(X_\text{IMG} + E_\text{pos} | k, s) W_q
\label{eq:proj_query}
\end{equation}
are down-sampled from the image sequence encoded with absolute position embeddings $E_\text{pos}$ by average pooling $\mathrm{Pool}(\cdot | k, s)$ with kernel size $k$ and stride $s$\footnotemark, and keys
\begin{equation}
    X_k = D_k(X_\text{IMG} + E_\text{pos}) = (X_\text{IMG} + E_\text{pos}) W_k
\label{eq:proj_key}
\end{equation}
are linearly projected from $X_\text{IMG}$ for computational efficiency (by setting $d_c \ll d$).
Finally, the computation of compression matrix $P$ is formulated as:
\begin{equation}
    P = \sigma\left(\frac{\tilde{X}_q \ \mathrm{diag}(\omega) X_k^\top}{\sqrt{d_c}}\right) \text{,}
\label{eq:comp_matrix}
\end{equation}
where diagonal matrix $\omega\in\mathbb{R}^{d_c}$ is learnable.

\footnotetext{\Cref{app:implementation} provides the detailed implementation for down-sampling $X_\text{IMG}$ to arbitrary length $m$.}

Regarding the module placement, following the setting of LLaVA-PruMerge~\cite{shang2024llava}, we apply our reduction module only before the LLM decoder, although \method can in principle be inserted at any layer.
This placement reduces the additional MHA computation incurred in earlier LLM layers compared with inserting it at intermediate layers, which is particularly beneficial for long visual inputs such as videos.
Moreover, our work focuses on developing a learning-based reduction method rather than a full-stage compression strategy across both the vision tower and the LLM, as explored in IVTP~\cite{kai2025ivtp}. Such full-stage optimization requires costly pretraining and instruction-tuning, which we leave as future work under our lightweight framework.

For the module design, since our primary goal is to reduce the inference burden of LVLMs, we intentionally avoid constructing complex reduction modules, such as those required for auto-regressive generation, as they would significantly increase latency and reduce computational efficiency. As the first learning-based token reduction framework, there are currently few non-learning approaches available for direct comparison. Nevertheless, we further discuss the relationship between our method and other data-driven approaches for efficient model inference in~\Cref{sec:discussions}.

\subsection{Theoretical Analysis}
\label{sec:method-theory}
Now we provide a theoretical analysis of \method to explain the design motivation and further introduce the optimization objectives and constraints.
To begin with, we expand~\Cref{eq:comp_matrix} as
\begin{equation}
    P_\text{raw} = \mathrm{Pool}(X|k,s) W_q \ \mathrm{diag}(\omega) W_k^\top X^\top \text{.}
\label{eq:comp_matrix_raw}
\end{equation}
Further, suppose all elements in $W$ are drawn independently from a Gaussian distribution $\mathcal{N}(0, \sigma_c^2)$ and with a specific initialization (\ie, $W_q = W_k$), \method will initially behave as a weighted pooling of the input image sequence (controlled by the kernel size $k$), as we prove in~\Cref{app:implementation-properties}.
Moreover, the meta generator will learn, in a data-driven manner, how to select and merge the visual tokens to minimize the prediction discrepancy.
Since we do not rely on any annotation for the compression matrix, the low-rank positive semi-definite form presented in \Cref{eq:comp_matrix_raw} provides a good starting point and facilities gradient decent optimization.

\subsection{Training \method} \label{sec:method-training}

\setlength{\textfloatsep}{6pt}
\begin{algorithm}[!t]
\caption{Training algorithm for \method.}
\label{alg:comp}
\begin{algorithmic}[1]
    \REQUIRE $E_\text{TXT}(\cdot)$: the language encoder;
    $V_\text{IMG}(\cdot)$: the image encoder;
    $\mathrm{LLM}(\cdot, \cdot)$: the vision-language decoder;
    $\mathcal{P}_\text{meta}(\cdot | \Theta)$: the proposed \method module with learnable parameters $\Theta$.
    \FOR{$(I_\text{IMG}, I_\text{TXT}) \in \mathcal{D}_\text{train}$}
        \STATE $X_\text{TXT} \gets E_\text{TXT}(I_\text{TXT})$
        \STATE $X_\text{IMG} \gets V_\text{IMG}(I_\text{IMG})$
        \STATE $\tilde{X}_\text{IMG} \gets \mathcal{P}_\text{meta}(X_\text{IMG} | \Theta)$
        \COMMENT{with gradients}
        \STATE $\bm{y} \gets \mathrm{LLM}(X_\text{TXT}, X_\text{IMG})$
        \STATE $\bm{\tilde{y}} \gets \mathrm{LLM}(X_\text{TXT}, \tilde{X}_\text{IMG})$
        \COMMENT{with gradients}
        \STATE Compute the final loss and gradient $\nabla_\Theta$ \wrt $\Theta$.
        \STATE Update $\Theta$ with SGD optimizer.
    \ENDFOR
\end{algorithmic}
\end{algorithm}

Similar to the training objective introduced in~\Cref{sec:method-optimal_mapping}, we train \method by minimizing the prediction discrepancy $\mathcal{L}_\text{pred}$ with additional sparsity regularization $\mathcal{L}_\text{entropy}$.
However, due to the lack of ground-truth compression matrix, the generated compression matrix $P$ tends to collapse to trivial solutions where the compressed tokens all derive from the same input source.
To avoid this, we add a collapse regularization term $\mathcal{L}_\text{collapse} = \max_{j} \sum_{i=1}^m P_{i,j}$.
Therefore, the final optimization objective is
\begin{equation}
    \mathcal{L} = \mathcal{L}_{pred} + \alpha_\text{entropy} \mathcal{L}_\text{entropy} + \alpha_\text{collapse} \mathcal{L}_\text{collapse} \text{,}
\label{eq:loss}
\end{equation}
where $\alpha_\text{entropy}$ and $\alpha_\text{collapse}$ are hyperparameters.
\Cref{alg:comp} delineates the training procedure.

    \begin{table*}[!t]
\centering
\setlength{\tabcolsep}{5pt}
\caption{The comparison of visual token reduction methods on three MT-VQA benchmarks with the reduction rate of 90\%. The best and the second-best results are highlighted in bold and underline, respectively.}
\resizebox{\linewidth}{!}{%
\begin{tabular}{ll @{\hspace{15pt}} *{4}{r} @{\hspace{15pt}} *{4}{r} @{\hspace{15pt}} *{4}{r}}
    \toprule
    \multicolumn{1}{c}{\multirow{2}{*}{\textbf{Model}}}
    & \multicolumn{1}{c@{\hspace{15pt}}}{\multirow{2}{*}{\textbf{Method}}}
    & \multicolumn{4}{c}{\textbf{MT-VQA-v2}}
    & \multicolumn{4}{c}{\textbf{MT-GQA}}
    & \multicolumn{4}{c}{\textbf{ConvBench}} \\
    \cmidrule(lr){3-6} \cmidrule(lr){7-10}  \cmidrule(lr){11-14}
    & & \multicolumn{1}{c}{$Acc_1$}
    & \multicolumn{1}{c}{$Acc_2$}
    & \multicolumn{1}{c}{$Acc_3$}
    & \multicolumn{1}{c@{\hspace{15pt}}}{$Avg$}
    & \multicolumn{1}{c}{$Acc_1$}
    & \multicolumn{1}{c}{$Acc_2$}
    & \multicolumn{1}{c}{$Acc_3$}
    & \multicolumn{1}{c@{\hspace{15pt}}}{$Avg$}
    & \multicolumn{1}{c}{$S_1$}
    & \multicolumn{1}{c}{$S_2$}
    & \multicolumn{1}{c}{$S_3$}
    & \multicolumn{1}{c}{$Avg$} \\
    \midrule

    \multirow{6}{*}{LLaVA-1.5-7b} & \textcolor{gray}{Base}
    & \textcolor{gray}{76.72} & \textcolor{gray}{77.51} & \textcolor{gray}{77.30} & \textcolor{gray}{77.18} & \textcolor{gray}{61.76} & \textcolor{gray}{64.07} & \textcolor{gray}{65.35} & \textcolor{gray}{63.73} & \textcolor{gray}{4.33} & \textcolor{gray}{5.72} & \textcolor{gray}{5.55} & \textcolor{gray}{5.20} \\
    & Random
    & 66.36 & 66.94 & 66.68 & 66.66 & 54.60 & 57.07 & 59.31 & 56.99 & 3.73 & 3.99 & 4.08 & 3.93 \\
    & Sample
    & 67.11 & 67.52 & 67.63 & 67.42 & 55.06 & 57.89 & 59.74 & \uline{57.56} & 3.64 & 4.85 & 3.81 & \uline{4.10} \\
    & FastV
    & 45.98 & 48.56 & 49.65 & 48.06 & 40.98 & 46.71 & 49.30 & 45.66 & 1.56 & 1.39 & 3.12 & 2.02 \\
    & PruMerge
    & 69.03 & 69.93 & 69.73 & \uline{69.56} & 55.26 & 57.23 & 60.13 & 57.54 & 4.51 & 3.47 & 3.47 & 3.82 \\
    & Ours
    & 70.27 & 70.31 & 71.36 & \textbf{70.65} & 55.95 & 58.71 & 60.64 & \textbf{58.43} & 4.33 & 3.99 & 4.16 & \textbf{4.16} \\
    \midrule

    \multirow{6}{*}{LLaVA-1.5-13b}  & \textcolor{gray}{Base}
    & \textcolor{gray}{78.35} & \textcolor{gray}{79.47} & \textcolor{gray}{78.92} & \textcolor{gray}{78.91} & \textcolor{gray}{62.47} & \textcolor{gray}{65.21} & \textcolor{gray}{67.22} & \textcolor{gray}{64.97} & \textcolor{gray}{4.33} & \textcolor{gray}{7.11} & \textcolor{gray}{5.72} & \textcolor{gray}{5.72} \\
    & Random
    & 67.26 & 68.05 & 67.63 & 67.65 & 54.63 & 57.87 & 60.23 & 57.58 & 3.56 & 5.55 & 4.42 & 4.51 \\
    & Sample
    & 68.12 & 68.82 & 68.47 & 68.47 & 55.41 & 58.53 & 60.26 & 58.07 & 4.16 & 5.03 & 4.33 & 4.51 \\
    & FastV
    & 55.36 & 56.80 & 57.08 & 56.41 & 49.08 & 53.34 & 56.14 & 52.85 & 2.19 & 3.47 & 4.20 & 3.29 \\
    & PruMerge
    & 70.18 & 71.16 & 70.70 & \uline{70.68} & 55.70 & 57.94 & 60.70 & \uline{58.11} & 4.51 & 3.99 & 5.55 & \uline{4.68} \\
    & Ours
    & 72.70 & 73.24 & 72.88 & \textbf{72.94} & 57.02 & 59.26 & 62.16 & \textbf{59.48} & 4.68 & 4.51 & 6.41 & \textbf{5.20} \\

    \midrule
    \multirow{5}{*}{LLaVA-NeXT-7b}  & \textcolor{gray}{Base}
    & \textcolor{gray}{80.20} & \textcolor{gray}{80.86} & \textcolor{gray}{80.71} & \textcolor{gray}{80.59} & \textcolor{gray}{63.83} & \textcolor{gray}{66.68} & \textcolor{gray}{67.94} & \textcolor{gray}{66.15} & \textcolor{gray}{7.95} & \textcolor{gray}{11.46} & \textcolor{gray}{7.58} & \textcolor{gray}{9.00} \\
    & Random
    & 70.65 & 72.26 & 72.44 & 71.78 & 58.60 & 61.04 & 63.00 & 60.88 & 5.81 & 6.59 & 4.42 & \uline{5.61} \\
    & Sample
    & 70.88 & 72.32 & 72.35 & \uline{71.85} & 58.46 & 61.39 & 63.24 & \uline{61.03} & 3.81 & 7.28 & 5.72 & 5.60 \\
    & FastV
    & 57.09 & 59.00 & 59.27 & 58.45 & 46.42 & 50.55 & 53.95 & 50.31 & 0.00 & 1.85 & 1.85 & 1.23 \\
    & Ours
    & 73.83 & 75.24 & 75.18 & \textbf{75.18} & 59.43 & 63.49 & 65.19 & \textbf{62.70} & 4.16 & 8.67 & 9.01 & \textbf{7.28} \\

    \midrule
    \multirow{5}{*}{LLaVA-NeXT-13b} & \textcolor{gray}{Base}
    & \textcolor{gray}{81.02} & \textcolor{gray}{82.32} & \textcolor{gray}{81.64} & \textcolor{gray}{81.66} & \textcolor{gray}{65.45} & \textcolor{gray}{67.32} & \textcolor{gray}{69.12} & \textcolor{gray}{67.30} & \textcolor{gray}{12.48} & \textcolor{gray}{13.17} & \textcolor{gray}{7.97} & \textcolor{gray}{11.21} \\
    & Random
    & 71.86 & 73.44 & 73.22 & 72.84 & 59.61 & 61.86 & 63.43 & 61.63 & 7.20 & 9.19 & 6.33 & 7.57 \\
    & Sample
    & 71.97 & 73.84 & 73.43 & \uline{73.08} & 59.69 & 62.28 & 63.88 & \uline{61.95} & 6.07 & 10.92 & 6.07 & \uline{7.69} \\
    & FastV
    & 57.07 & 59.09 & 59.14 & 58.43 & 47.23 & 51.34 & 53.36 & 50.64 & 5.17 & 5.17 & 1.72 & 4.02 \\
    & Ours
    & 74.62 & 75.73 & 75.42 & \textbf{75.26} & 60.78 & 63.41 & 65.16 & \textbf{63.12} & 6.93 & 11.27 & 6.76 & \textbf{8.32} \\
    
    \midrule
    \multirow{5}{*}{XComposer-2.5-7b}  & \textcolor{gray}{Base}
    & \textcolor{gray}{78.80} & \textcolor{gray}{81.12} & \textcolor{gray}{81.24} & \textcolor{gray}{80.39} & \textcolor{gray}{60.21} & \textcolor{gray}{63.38} & \textcolor{gray}{65.01} & \textcolor{gray}{62.87} & \textcolor{gray}{12.48} & \textcolor{gray}{7.00} & \textcolor{gray}{7.97} & \textcolor{gray}{9.15} \\
    & Random
    & 67.88 & 70.20 & 70.49 & 69.52 & 52.52 & 57.01 & 60.67 & 56.73 & 11.35 & 9.88 & 7.28 & 9.50 \\
    & Sample
    & 68.46 & 70.92 & 70.78 & 70.05 & 52.94 & 57.20 & 59.69 & 56.61 & 12.13 & 9.53 & 7.63 & \uline{9.76} \\
    & FastV
    & 72.65 & 75.02 & 75.02 & \uline{74.23} & 54.84 & 57.33 & 58.83 & \uline{57.00} & 4.17 & 4.17 & 0.00 & 2.78 \\
    & Ours
    & 73.91 & 76.68 & 76.70 & \textbf{75.76} & 55.99 & 58.49 & 61.55 & \textbf{58.68} & 12.31 & 9.71 & 7.63 & \textbf{9.88} \\

    \bottomrule
\end{tabular}%
}%
\label{table:result-comp-0.1}
\end{table*}

\section{Experiments} \label{sec:exp}

\subsection{Implementation}
\noindent \textbf{Datasets.} We evaluate our method on three MT-VQA benchmarks: MT-VQA-v2, MT-GQA, and ConvBench\footnotemark.
MT-VQA-v2 is constructed based on the validation set of VQA-v2~\cite{antol2015vqa,zhang2016yin} with 25k three-turn image-dialogue pairs.
Similarly, MT-GQA is constructed from the testdev-balanced set of GQA~\cite{hudson2019gqa} with 4061 three-turn dialogues.
\footnotetext{As ConvBench relies on GPT-3.5-turbo’s commercial API for evaluation, we replace it with the recently released open-source LVLM, Llama-3.1-8B-Instruct~\cite{dubey2024llama}.}
ConvBench~\cite{liu2024convbench} is a native multi-turn conversation evaluation benchmark with 577 conversations that adopts a three-level multimodal capability hierarchy.
Instead of training on the entire dataset, which is time-consuming, we only train \method on a small subset (about 20k items) drawn from the training-balanced split of MT-GQA and the training set of MT-VQA-v2.
We utilize the pre-trained weights on MT-VQA-v2 to evaluate on ConvBench.

\noindent \textbf{LVLMs.} To evaluate the generalizability of our method, we choose five different LVLMs: LLaVA-1.5-7b/13b~\cite{liu2023improvedllava}, LLaVA-NeXT-7b/13b~\cite{liu2024llavanext}, and InternLM-XComposer-2.5-7b~\cite{internlmxcomposer2_5}.
Of these models, LLaVA-1.5 employs a single-scale vision tower with a fixed visual sequence length, while the others adopt multi-scale perception, resulting in variable visual sequence lengths, which brings further challenges to the token reduction method.

\noindent \textbf{Training Details and Selection of Hyperparameters.}
We implement our method with PyTorch~\cite{paszke2019pytorch} and optimize the proposed \method with SGD~\cite{ruder2016overview} with a learning rate of $10^{-3}$.
Gradient clipping is adopted with a maximum value of $10^{-2}$.
We train all the settings for 2 epochs with a batch size of 36 on four commercial NVIDIA RTX A6000 GPUs.
The training of LLaVA-NeXT-7B with a 90\% reduction rate takes approximately 30 GPU hours, which corresponds to about only 9 hours on a 4-GPU machine.
We initialize $W_q=W_k$ and drawn from Gaussian distribution $\mathcal{N}(0, \frac{1}{\sqrt{d_c}}^2)$; $\omega$ is set to all ones; $\alpha_\text{entropy} = \alpha_\text{collapse}=1$ as the default setting.

\begin{figure*}[!t]
\centering%
\begin{subfigure}[b]{0.5\linewidth}
    \centering
    \includegraphics[width=0.9\linewidth]{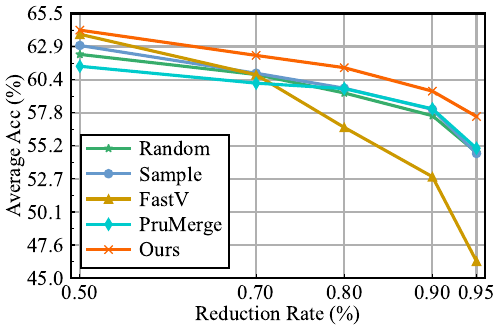}%
    \caption{LLaVA-1.5-13b}%
    \vspace{-5pt}
\label{fig:exp-comp_rate-llava_1.5}%
\end{subfigure}%
\begin{subfigure}[b]{0.5\linewidth}
    \centering
    \includegraphics[width=0.9\linewidth]{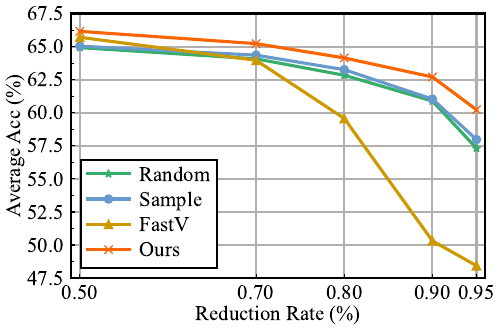}%
    \caption{LLaVA-NeXT-7b}%
    \vspace{-5pt}
\label{fig:exp-comp_rate-llava_1.6}%
\end{subfigure}%
\caption{Comparison of average accuracy on MT-GQA with reduction rate from 50\% to 95\%.}%
\label{fig:exp-comp_rate}%
\end{figure*}%

\subsection{Comparison Results} \label{sec:exp-main}

We evaluate the proposed \method and comparison baselines with the following settings:
\begin{itemize}
    \item \textbf{Base:} The base LVLM evaluated directly on the MT-VQA benchmarks without token reduction.
    \item \textbf{Random:} Randomly prune visual tokens before the first layer of the LLM decoder. We report the average performance over 3 random seeds.
    \item \textbf{Sample:} Perform equidistant down-sampling on the visual sequence before the LLM decoder.
    \item \textbf{FastV:} Our implementation of FastV~\cite{dosovitskiy2021an} for multi-scale vision tower. The guidance attention weights are extracted from the first layer of the LLM decoder and visual tokens are pruned at the second layer.
    \item \textbf{PruMerge:} Perform LLaVA-PruMerge~\cite{liu2023improvedllava} only for LLaVA-1.5, as it is not compatible with the multi-scale visual tower.
    \item \textbf{Ours:} Perform our proposed \method before the LLM decoder.
\end{itemize}

The comparison results with reduction rate 90\% and 70\% are shown in~\Cref{table:result-comp-0.1} and~\Cref{table:result-comp-0.3}, where we compare the accuracy of each turn conversation and the overall accuracy on three MT-VQA benchmarks.
\emph{It is noticeable that the proposed \method consistently outperforms the baseline methods.}
While not trained on ConvBench, our method still surpasses the baseline methods by a large margin, demonstrating the transferability of \method.
For LLaVA-1.5, experimental results show that LLaVA-PruMerge which is designed specifically for it performs slightly better than Sample, but still lags behind our approach.
On the other hand, FastV performs significantly worse than both the Sample and even the Random methods.
This further supports our findings in~\Cref{sec:method-optimal_mapping}, where we have revealed that using attention as guidance for compression results in a loss of critical tokens.
Although FastV shows some improvement for XComposer-2.5-7b, it still performs poorly on ConvBench.

\Cref{fig:exp-comp_rate} illustrates the performance curve of different methods with varying reduction rates from 50\% to 95\%.
It is clear that our method consistently outperforms the baselines across different reduction rates, while FastV performers better for low reduction rate and LLaVA-PruMerge performers better for high reduction rate.

\begin{table}[!t]
\centering
\setlength{\tabcolsep}{3pt}
\caption{
Efficiency comparison of different token reduction methods.
The time to first token~(TTFT, ms), end-to-end generation time~(E2ET, ms), GPU memory usage~(Mem. GB), and TFLOPs are reported on MT-GQA dataset with a reduction rate of 90\%.
}
\vspace{-0.2em}
\resizebox{\linewidth}{!}{%
\begin{tabular}{ll *{4}{c}}
    \toprule
    \multicolumn{1}{c}{\textbf{Model}}
    & \multicolumn{1}{c}{\textbf{Setting}}
    & \multicolumn{1}{c}{\textbf{TTFT}}
    & \multicolumn{1}{c}{\textbf{E2ET}}
    & \multicolumn{1}{c}{\textbf{Mem.}}
    & \multicolumn{1}{c}{\textbf{TFLOPs}}\\
    \midrule
    \multirow{6}{*}{LLaVA-1.5-7b}
    & Base
    & 232 {\color{gray}\scriptsize($\pm$ 5.8)} & 676 {\color{gray}\scriptsize($\pm$ 8.0)} & 26.9 & 71.4 \\
    & Random
    & 98.2 {\color{gray}\scriptsize($\pm$ 5.7)} & 487 {\color{gray}\scriptsize($\pm$ 4.6)} & 26.2 & 13.3 \\
    & Sample
    & 96.9 {\color{gray}\scriptsize($\pm$ 5.8)} & 482 {\color{gray}\scriptsize($\pm$ 4.9)} & 26.2 & 13.3 \\
    & FastV
    & 102 {\color{gray}\scriptsize($\pm$ 4.52)} & 528 {\color{gray}\scriptsize($\pm$ 6.0)} & 26.3 & 13.5 \\
    & PruMerge
    & 107 {\color{gray}\scriptsize($\pm$ 5.23)} & 509 {\color{gray}\scriptsize($\pm$ 4.5)} & 26.2 & 13.3 \\
    & Ours
    & 97.8 {\color{gray}\scriptsize($\pm$ 5.41)} & 480 {\color{gray}\scriptsize($\pm$ 5.1)} & 26.1 & 13.3 \\
 
    \midrule
    \multirow{6}{*}{LLaVA-NeXT-7b}
    & Base
    & 484 {\color{gray}\scriptsize($\pm$ 4.7)} & 830 {\color{gray}\scriptsize($\pm$ 13.5)} & 16.7 & 95.3 \\
    & Random
    & 174 {\color{gray}\scriptsize($\pm$ 3.4)} & 481 {\color{gray}\scriptsize($\pm$ 4.5)} & 14.8 & 12.7 \\
    & Sample
    & 176 {\color{gray}\scriptsize($\pm$ 3.2)} & 484 {\color{gray}\scriptsize($\pm$ 5.3)} & 14.8 & 12.7 \\
    & FastV
    & 219 {\color{gray}\scriptsize($\pm$ 5.0)} & 529 {\color{gray}\scriptsize($\pm$ 5.33)} & 19.2 & 12.9 \\
    & Ours
    & 174 {\color{gray}\scriptsize($\pm$ 6.1)} & 501 {\color{gray}\scriptsize($\pm$ 4.8)} & 14.9 & 12.7 \\
    \bottomrule
\end{tabular}%
}
\label{tab:efficiency}
\end{table}
\subsection{Efficiency Results} \label{sec:exp-efficiency}
\Cref{tab:efficiency} compares the inference efficiency of different token reduction methods.
As we can observe that our method achieves compatible efficiency with the `Sample' setting, which is the most efficient baseline thanks to the explicit low-ranking mechanism as described in~\Cref{eq:comp_matrix_raw}.

\subsection{Transfer Results} \label{sec:exp-transfer}
Beyond the transfer results on ConvBench in~\Cref{table:result-comp-0.1}, we further evaluate the transfer capability of \method through comprehensive cross-dataset validation. Specifically, we perform transfer learning experiments between MT-GQA and MT-VQA-V2, with the results summarized in~\Cref{tab:transfer}. This table reports the average accuracy under a 90\% token reduction rate for both directions of transfer, from MT-GQA to MT-VQA-V2 and vice versa. These results indicate that \method is not heavily dependent on a specific training dataset, demonstrating robust generalization. We also conduct transfer experiments on the video question answering task, as reported in~\Cref{tab:video} of~\Cref{app:transfer}.

\vspace{-0.1em}
\subsection{Ablation Study} \label{sec:exp-ablation}
\vspace{-0.1em}

\begin{table}[!t]
\centering
\setlength{\tabcolsep}{8pt}
\caption{Ablation study of training \method for LLaVA-NeXT-7b using different loss terms on MT-GQA. Gradient clipping is only applied for the `$\mathcal{L}_\text{collapse}$ + \textit{Grad Clip}' setting.}
\resizebox{\linewidth}{!}{%
\begin{tabular}{*{7}{c}}
    \toprule
    \multicolumn{4}{c}{\textbf{Settings}}
    & \multicolumn{1}{c}{\textbf{MT-GQA}} \\
    \cmidrule(lr){1-4}
    \cmidrule(lr){5-5}
    \multicolumn{1}{c}{$\mathcal{L}_\text{pred}$}
    & \multicolumn{1}{c}{$\mathcal{L}_\text{entropy}$}
    & \multicolumn{1}{c}{$\mathcal{L}_\text{collapse}$}
    & \multicolumn{1}{c}{$\mathcal{L}_\text{collapse}$ + \textit{Grad Clip}}
    & \multicolumn{1}{c}{$Avg$} \\
    \midrule
    \cmark & \xmark & \xmark & \xmark & 61.98 \\
    \cmark & \cmark & \xmark & \xmark & 62.42 \\
    \cmark & \xmark & \cmark & \xmark & 56.34 \\
    \cmark & \xmark & \xmark & \cmark & 62.13 \\
    \midrule
    \cmark & \cmark & \xmark & \cmark & 62.70 \\
    \bottomrule
\end{tabular}%
}%
\label{tab:ablation}
\end{table}

As delineated in~\Cref{sec:method-training}, we utilize three optimization objectives to train the proposed method.
To investigate the effectiveness of each objective, we conduct an ablation study by removing one of the objectives at a time.
The results in~\Cref{tab:ablation} (with additional results for various LVLMs in~\Cref{tab:ablation_more}) demonstrate that each objective contributes positively to the overall performance.
In particular, training utilizing the $\mathcal{L}_\text{collapse}$ alone leads to divergence because of the relatively high penalty on the collapse objective, especially when the reduction rate is small (less than 70\%).
To tackle this, we introduce gradient clipping to stabilize the training process.

\begin{figure}[!h]
\centering%
\includegraphics[width=0.95\linewidth]{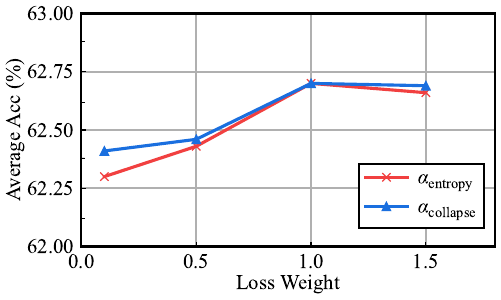}%
\caption{Sensitivity analysis in training \method for LLaVA-NeXT-7b with different weights $\alpha_\text{entropy}$ and $\alpha_\text{collapse}$ on MT-GQA.}%
\label{fig:exp-sens}%
\end{figure}

Besides, we also investigate the sensitivity of the hyperparameters $\alpha_\text{entropy}$ and $\alpha_\text{collapse}$ when training \method.
\Cref{fig:exp-sens} shows the performance curves for different weight settings, demonstrating that the performance remains relatively stable (within a 0.5 percentage point variation).

\vspace{-0.1em}
\subsection{Visualization}  \label{sec:exp-vis}
\begin{figure}[!t]
\centering%
\begin{subfigure}{\linewidth}%
\centering
\includegraphics[width=\textwidth]{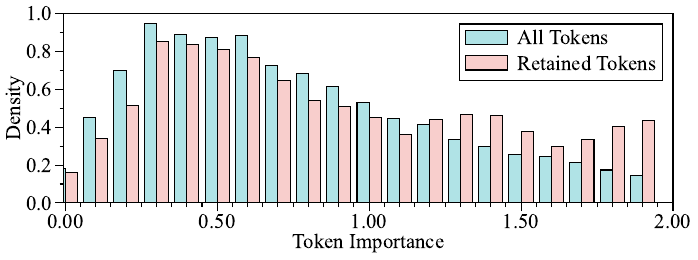}%
\vspace{-0.5em}%
\caption{}%
\end{subfigure} \\
\begin{subfigure}{\linewidth}%
\centering
\includegraphics[width=\textwidth]{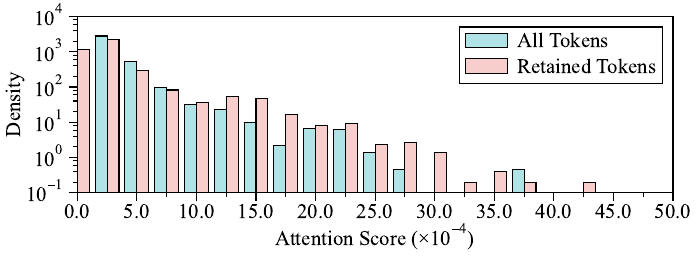}%
\vspace{-0.5em}%
\caption{}%
\end{subfigure}
\caption{(a) Token importance distribution. (b) Attention distribution over prompt tokens. Image tokens are extracted from the last layer of the vision tower of LLaVA-NeXT-7b on VQA-v2 dataset.}%
\label{fig:vis}%
\end{figure}

\vspace{-0.1em}
\Cref{fig:vis} visualizes the attention distribution for LLaVA-NeXT-7b, similar to~\Cref{fig:comp_one_img}.
As directly computing the attention to \cls token is not feasible for multi-scale vision towers, we compute FastV's style image token importance instead.
Nevertheless, we observe that only a small number of tokens with high attention are retained, which is consistent with the conclusion in~\Cref{sec:method-optimal_mapping} and further demonstrates that using token attention to guide reduction is suboptimal.

    \vspace{-0.3em}
    \section{Conclusion and Outlook} \label{sec:conclusion}
\vspace{-0.1em}

This paper proposes a novel token reduction approach for multi-turn VQA scenarios.
To this end, we first unify token pruning and merging under the framework of compression projection to visual sequences and explore the optimal compression mapping for a single image.
Preliminary results reveal that existing methods guided by attention are suboptimal, as a large number of retained tokens do not correspond to the highest attention scores.
This motivates us to further explore the construction of an optimal compression mapping for the entire dataset.
To achieve this, we propose \method, a meta generator conditioned solely on the visual sequence, and optimized in a data-driven manner.
Extensive experiments demonstrate the efficiency and effectiveness of our method.
In future work, we will explore the token reduction strategy for all LLM layers without a hand-crafted design, and investigate the transferability of our method to more challenging tasks, such as video understanding.

\section*{Acknowledgments}
This work is funded by National Natural Science Foundation of China (62576305), the Alibaba Group through Alibaba Innovative Research Program, and the National Research Foundation, Singapore, under its Medium Sized Center for Advanced Robotics Technology Innovation.

\clearpage

    {%
        \small
        \bibliography{%
            references/header.bib,
            references/efficiency.bib,
            references/others.bib,
            references/models.bib
        }
        \bibliographystyle{ieeenat_fullname}
    }

\clearpage
\setcounter{page}{1}
\maketitlesupplementary

\section{Implementation Details} \label{app:details}

\subsection{Fixed Compression Matrix} \label{app:optimial_compression-algorithm}
The training algorithm for $P_\text{raw}$ is shown in~\Cref{alg:train_p}, where compression projection matrix $P_\text{raw}$ is optimized to minimize the objective as described in~\Cref{eq:loss_single_img}.

There are several hyperparameters involved in the training of $P_\text{raw}$, including $\alpha$ in~\Cref{eq:loss_single_img} and $\sigma_\text{raw}$ in the initialization of $P_\text{raw}$.
We set $\alpha = 1$ and $\sigma_\text{raw} = 0.1$ in all experiments.
The learning rate is set to $10$ and we train 500 epochs for each image-text pair.

\begin{algorithm}[h]
\caption{Training algorithm for \method.}
\label{alg:train_p}
\begin{algorithmic}[1]
    \REQUIRE $(I_\text{IMG}, I_\text{TXT})$: the image-text pair; $E_\text{TXT}(\cdot)$: the language encoder; $V_\text{IMG}(\cdot)$: the image encoder; $\mathrm{LLM}(\cdot, \cdot)$: the vision-language decoder; $P_\text{raw}$: the learnable compression matrix with shape $m\times n$.
    \STATE Initialize $P_\text{raw}$ with Gaussian distribution $\mathcal{N}(0, \sigma_\text{raw}^2)$.
    \STATE $X_\text{TXT} \gets E_\text{TXT}(I_\text{TXT})$
    \STATE $X_\text{IMG} \gets V_\text{IMG}(I_\text{IMG})$
    \STATE $\bm{y} \gets \mathrm{LLM}(X_\text{TXT}, X_\text{IMG})$
    \WHILE{\textit{not} converged}
        \STATE $\tilde{X}_\text{IMG} \gets \sigma(P_\text{raw})X_\text{IMG}$
        \COMMENT{with gradients}
        \STATE $\bm{\tilde{y}} \gets \mathrm{LLM}(X_\text{TXT}, \tilde{X}_\text{IMG})$
        \COMMENT{with gradients}
        \STATE Compute the final loss and gradient $\nabla_{P_\text{raw}}$ \wrt $P_\text{raw}$.
        \STATE Update $P_\text{raw}$ with SGD optimizer.
    \ENDWHILE
\end{algorithmic}
\end{algorithm}

\subsection{\method} \label{app:implementation}
To adapt to arbitrary compression rate, the stride $k$ in the pooling operation is set to a float value $\frac{n}{m}$, which can be easily implemented by the fractional max pooling operation~\cite{graham2014fractional}.
We set the kernel size $s = 3$ for all experiments.

\section{Properties of \method} \label{app:implementation-properties}
Now, we analyze the properties of \method when $W_q = W_k = W$ and are drawn from $\mathcal{N}(0, \sigma_w^2)$ in~\Cref{eq:comp_matrix_raw}.

We start by considering when kernel size $k = 1$, meaning that $\mathrm{Pool}(X)$ is a down-sampling operation to $X$, and we let $\tilde{X}$ denotes the down-sampled image sequence.
In this case, \Cref{eq:comp_matrix_raw} can be simplified as
\begin{equation}
    P = \tilde{X} S X^\top \text{,}
\end{equation}
where $S$ is a positive semi-definite matrix.
Therefore, the expectation $\mathbb{E}[p_{i,j}] = 0$ for all positions that $\tilde{x}_i \neq x_j$, and for $\tilde{x}_i = x_j = x$
\begin{equation}
    \mathbb{E}[p_{i,j}] = \mathbb{E}[xSx^\top]  = \mathbb{E}[xW\ \mathrm{diag}(\omega)W^\top x^\top] \text{.}
\end{equation}
Here, we notice that $y = xW$ is still a random vector, with all elements subject to $\mathcal{N}(0, d\sigma_c^2 \sigma_w^2)$.
Hence,
\begin{equation}
    \mathbb{E}[y \ \mathrm{diag}(\omega)y^\top] = d d_c  \sigma_c^2 \sigma_w^2 \text{.}
\end{equation}

Considering that the embedding dimension of LVLMs is a large number (\eg, 4096 for LLaVA-1.5-7b), $\sigma(P_\text{raw})$ is close to the down-sampling projection to the input image sequence controlled by the stride $s$.

Further, when the kernel size $k > 1$, the expectation of $p_{i,j}$ is still zero when $\tilde{x}_i$ is not captured by the pooling kernel located in $x_i$.
To sum up, the initialization to \Cref{eq:comp_matrix_raw} converting \method to a interpretable pooling operation to the input image sequence.

However, as training progresses, $W_q$ diverges from $W_k$, breaking the positive semi-definiteness of matrix $S$, enabling \method to further explore more effective compression strategies, ultimately enhancing performance.
Besides, we choose the compression embedding dimension $d_c$ to be smaller than the original embedding dimension $d$ to reduce the computational cost and number of parameters.

Essentially, \Cref{eq:comp_matrix_raw} is a specialized form of the dot-product attention $X W_Q W_K^\top X^\top$, making it easier to optimize and less prone to over-fitting to the training dataset, as we only adopt a few-shot subset for training efficiency.

\section{More Results} \label{app:more_results}

\subsection{Performance of Fixed Compression Matrix}
Because we train the compression matrix $P_\text{raw}$ for a single image on the training dataset, which is a straightforward optimization problem, we do not compare it with other methods.
Here, we present the overall accuracy about compressing LLaVA-Next-7b on the MT-VQA-v2 dataset for reference.
The accuracy of the base setting is 82.44, and when reducing 90\% of the image token the accuracy decreases to 80.89.

\subsection{More comparison Results}
\label{app:comparison}

\begin{table*}[!ht]
\centering
\setlength{\tabcolsep}{5pt}
\caption{The comparison of visual token reduction methods on three MT-VQA benchmarks with the reduction rate of 70\%. The best and the second-best results are highlighted in bold and underline, respectively.}
\resizebox{\linewidth}{!}{%
\begin{tabular}{ll *{4}{r} *{4}{r} *{4}{r}}
    \toprule
    \multicolumn{1}{c}{\multirow{2}{*}{\textbf{Model}}}
    & \multicolumn{1}{c}{\multirow{2}{*}{\textbf{Method}}}
    & \multicolumn{4}{c}{\textbf{MT-VQA-v2}}
    & \multicolumn{4}{c}{\textbf{MT-GQA}}
    & \multicolumn{4}{c}{\textbf{ConvBench}} \\
    \cmidrule(lr){3-6} \cmidrule(lr){7-10}  \cmidrule(lr){11-14}
    & & \multicolumn{1}{c}{$Acc_1$}
    & \multicolumn{1}{c}{$Acc_2$}
    & \multicolumn{1}{c}{$Acc_3$}
    & \multicolumn{1}{c}{$Avg$}
    & \multicolumn{1}{c}{$Acc_1$}
    & \multicolumn{1}{c}{$Acc_2$}
    & \multicolumn{1}{c}{$Acc_3$}
    & \multicolumn{1}{c}{$Avg$}
    & \multicolumn{1}{c}{$S_1$}
    & \multicolumn{1}{c}{$S_2$}
    & \multicolumn{1}{c}{$S_3$}
    & \multicolumn{1}{c}{$Avg$} \\
    \midrule

    \multirow{6}{*}{LLaVA-1.5-7b} & \textcolor{gray}{Base}
    & \textcolor{gray}{76.72} & \textcolor{gray}{77.51} & \textcolor{gray}{77.30} & \textcolor{gray}{77.18} & \textcolor{gray}{61.76} & \textcolor{gray}{64.07} & \textcolor{gray}{65.35} & \textcolor{gray}{63.73} & \textcolor{gray}{4.33} & \textcolor{gray}{5.72} & \textcolor{gray}{5.55} & \textcolor{gray}{5.20} \\
    & Random
    & 72.72 & 73.57 & 73.11 & 73.13 & 57.79 & 61.07 & 63.09 & \uline{60.65} & 4.17 & 4.73 & 5.51 & \textbf{4.80} \\
    & Sample
    & 72.96 & 73.71 & 73.33 & 73.33 & 58.48 & 61.12 & 62.30 & 60.63 & 4.16 & 5.03 & 3.81 & 4.33 \\
    & FastV
    & 69.30 & 69.57 & 69.41 & 69.43 & 54.79 & 57.65 & 60.03 & 57.49 & 3.99 & 5.03 & 3.47 & 4.16 \\
    & PruMerge
    & 72.79 & 73.90 & 73.42 & \uline{73.37} & 57.89 & 59.54 & 61.81 & 59.75 & 3.64 & 4.68 & 4.68 & 4.33 \\
    & Ours
    & 75.67 & 76.63 & 76.46 & \textbf{76.25} & 58.62 & 60.96 & 63.64 & \textbf{61.07} & 3.99 & 5.03 & 4.85 & \uline{4.62} \\
    \midrule

    \multirow{6}{*}{LLaVA-1.5-13b}  & \textcolor{gray}{Base}
    & \textcolor{gray}{78.35} & \textcolor{gray}{79.47} & \textcolor{gray}{78.92} & \textcolor{gray}{78.91} & \textcolor{gray}{62.47} & \textcolor{gray}{65.21} & \textcolor{gray}{67.22} & \textcolor{gray}{64.97} & \textcolor{gray}{4.33} & \textcolor{gray}{7.11} & \textcolor{gray}{5.72} & \textcolor{gray}{5.72} \\
    & Random
    & 73.63 & 74.56 & 73.96 & 74.05 & 57.74 & 61.24 & 63.33 & 60.77 & 4.03 & 4.89 & 5.24 & 4.72 \\
    & Sample
    & 73.81 & 74.79 & 74.51 & 74.37 & 58.51 & 61.29 & 62.82 & \uline{60.87} & 3.64 & 5.03 & 5.37 & 4.68 \\
    & FastV
    & 73.85 & 75.18 & 74.58 & \uline{74.54} & 57.99 & 60.75 & 63.51 & 60.75 & 4.37 & 6.92 & 5.10 & \uline{5.46} \\
    & PruMerge
    & 73.80 & 75.16 & 74.57 & 74.51 & 57.67 & 60.45 & 62.18 & 60.10 & 4.51 & 6.41 & 5.37 & 5.43 \\
    & Ours
    & 74.03 & 76.98 & 76.31 & \textbf{75.77} & 59.48 & 61.91 & 65.25 & \textbf{62.21} & 4.33 & 6.93 & 5.37 & \textbf{5.55} \\

    \midrule
    \multirow{5}{*}{LLaVA-NeXT-7b}  & \textcolor{gray}{Base}
    & \textcolor{gray}{80.20} & \textcolor{gray}{80.86} & \textcolor{gray}{80.71} & \textcolor{gray}{80.59} & \textcolor{gray}{63.83} & \textcolor{gray}{66.68} & \textcolor{gray}{67.94} & \textcolor{gray}{66.15} & \textcolor{gray}{7.95} & \textcolor{gray}{11.46} & \textcolor{gray}{7.58} & \textcolor{gray}{9.00} \\
    & Random
    & 76.18 & 77.43 & 77.64 & 77.08 & 61.96 & 63.95 & 66.29 & 64.07 & 7.97 & 9.19 & 6.93 & \uline{8.03} \\
    & Sample
    & 76.60 & 77.93 & 77.96 & \uline{77.50} & 62.28 & 64.61 & 66.17 & \uline{64.35} & 7.63 & 8.32 & 4.33 & 6.76 \\
    & FastV
    & 75.96 & 76.86 & 76.39 & 76.40 & 61.54 & 64.37 & 65.97 & 63.96 & 0.00 & 0.00 & 2.50 & 0.83 \\
    & Ours
    & 77.75 & 78.06 & 78.54 & \textbf{78.12} & 63.38 & 64.69 & 67.59 & \textbf{65.22} & 7.63 & 9.88 & 7.45 & \textbf{8.32} \\

    \midrule
    \multirow{5}{*}{LLaVA-NeXT-13b} & \textcolor{gray}{Base}
    & \textcolor{gray}{81.02} & \textcolor{gray}{82.32} & \textcolor{gray}{81.64} & \textcolor{gray}{81.66} & \textcolor{gray}{65.45} & \textcolor{gray}{67.32} & \textcolor{gray}{69.12} & \textcolor{gray}{67.30} & \textcolor{gray}{12.48} & \textcolor{gray}{13.17} & \textcolor{gray}{7.97} & \textcolor{gray}{11.21} \\
    & Random
    & 77.30 & 78.77 & 78.65 & 78.24 & 62.89 & 64.64 & 67.22 & 64.92 & 11.44 & 11.27 & 8.15 & 10.29 \\
    & Sample
    & 77.51 & 79.15 & 79.03 & \uline{78.56} & 63.95 & 64.54 & 67.20 & \uline{65.23} & 10.40 & 14.04 & 6.76 & \uline{10.40} \\
    & FastV
    & 75.78 & 77.16 & 76.66 & 76.53 & 62.10 & 64.37 & 65.06 & 63.84 & 20.00 & 5.00 & 3.33 & 9.44 \\
    & Ours
    & 78.14 & 80.98 & 80.21 & \textbf{79.78} & 64.86 & 65.89 & 67.59 & \textbf{66.11} & 10.92 & 13.86 & 7.11 & \textbf{10.63} \\

    \midrule
    \multirow{5}{*}{XComposer-2.5-7b}  & \textcolor{gray}{Base}
    & \textcolor{gray}{78.80} & \textcolor{gray}{81.12} & \textcolor{gray}{81.24} & \textcolor{gray}{80.39} & \textcolor{gray}{60.21} & \textcolor{gray}{63.38} & \textcolor{gray}{65.01} & \textcolor{gray}{62.87} & \textcolor{gray}{12.48} & \textcolor{gray}{7.00} & \textcolor{gray}{7.97} & \textcolor{gray}{9.15} \\
    & Random
    & 74.63 & 77.23 & 77.44 & 76.43 & 56.96 & 61.59 & 63.63 & 60.73 & 15.42 & 10.23 & 7.97 & 11.21 \\
    & Sample
    & 75.07 & 77.68 & 78.04 & 76.93 & 57.79 & 61.17 & 63.36 & 60.77 & 15.94 & 11.79 & 6.07 & \uline{11.27} \\
    & FastV
    & 77.93 & 80.18 & 80.05 & \uline{79.39} & 58.95 & 61.34 & 62.67 & \uline{60.99} & 12.50 & 4.17 & 12.50 & 9.72 \\
    & Ours
    & 78.24 & 80.53 & 80.79 & \textbf{79.85} & 60.11 & 62.28 & 64.14 & \textbf{62.18} & 15.77 & 12.13 & 6.24 & \textbf{11.38} \\
    \bottomrule
\end{tabular}%
}%
\label{table:result-comp-0.3}
\end{table*}

\Cref{table:result-comp-0.3} presents the comparison results of different token reduction method with the compression rate of 70\%.
Our method achieves the best overall performance, the same as the results in \Cref{table:result-comp-0.1}.

\begin{table}[!ht]
\centering
\setlength{\tabcolsep}{13pt}
\caption{Comparison results of different token merging methods for LLaVA-1.5-7b.}%
\resizebox{\linewidth}{!}{%
\begin{tabular}{l rrrr}
    \toprule
    \multicolumn{1}{c}{\multirow{2}{*}{\textbf{Setting}}}
    & \multicolumn{4}{c}{\textbf{MT-GQA}} \\
    \cmidrule(lr){2-5}
    & \multicolumn{1}{c}{$Acc_1$}
    & \multicolumn{1}{c}{$Acc_2$}
    & \multicolumn{1}{c}{$Acc_3$}
    & \multicolumn{1}{c}{$Avg$} \\
    \midrule
    Base & 61.76 & 64.07 & 65.35 & 63.73 \\
    Sample & 54.60 & 57.07 & 59.31 & 56.99 \\
    Spatial & 51.05 & 54.62 & 56.24 & 53.97 \\
    ToMe & 53.64 & 56.91 & 57.62 &  56.06 \\
    VisionZip & 55.08 & 57.82 & 59.89 & 57.60 \\
    Ours & 55.95 & 58.71 & 60.64 & 58.43 \\
    \bottomrule
\end{tabular}%
}
\label{tab:merging}
\end{table}

As a supplement, \Cref{tab:merging} compares the effectiveness of token reduction methods.
Here, `Spatial' represents applying spatial pooling to the image sequence (the kernel size $k$ is set to the same as the stride $s$).
`ToMe'~\cite{bolya2023token} is a token merging method proposed for ViTs rather than LVLMs, and thus performs ineffectively in our setting.
`VisionZip'~\cite{Yang2024VisionZipLI} is a hybrid token compression method that integrates both token pruning and merging, yet it does not take MT-VQA scenarios into account and therefore also underperforms our method.

\subsection{More Ablation Results} \label{app:ablation}

\begin{table*}[!t]
\centering
\setlength{\tabcolsep}{12pt}
\caption{Additional ablation study of training \method for various LVLMs using different loss terms on MT-GQA. Gradient clipping is only applied for the `$\mathcal{L}_\text{collapse}$ + \textit{Grad Clip}' setting.}
\resizebox{\linewidth}{!}{%
\begin{tabular}{*{4}{c}ccc}
    \toprule
    $\mathcal{L}_\text{pred}$
    & $\mathcal{L}_\text{entropy}$
    & $\mathcal{L}_\text{collapse}$
    & $\mathcal{L}_\text{collapse}$ + \textit{Grad Clip}
    & \textbf{LLaVA-1.5-7b} 
    & \textbf{LLaVA-NeXT-7b} 
    & \textbf{XComposer-2.5-7b} \\
    \midrule
    \cmark & \xmark & \xmark & \xmark & 56.63 & 61.98 & 56.77 \\
    \cmark & \cmark & \xmark & \xmark & 57.99 & 62.42 & 58.01 \\
    \cmark & \xmark & \cmark & \xmark & 52.26 & 56.34 & 52.53 \\
    \cmark & \xmark & \xmark & \cmark & 57.57 & 62.13 & 58.24 \\
    \cmark & \cmark & \xmark & \cmark & 58.43 & 62.70 & 58.68 \\
    \bottomrule
\end{tabular}%
}
\label{tab:ablation_more}
\end{table*}
The results in~\Cref{tab:ablation_more} provide additional ablation studies across various LVLMs, further supporting the results in~\Cref{tab:ablation} and demonstrating that each objective contributes positively to the overall performance.

\subsection{More Transfer Results} \label{app:transfer}

\begin{table*}[!ht]
\centering
\setlength{\tabcolsep}{6pt}
\caption{Transfer validation experiments. Average accuracy is reported for cross-dataset transfer between MT-VQA-V2 and MT-GQA, all under a 90\% token reduction rate.}
\resizebox{\linewidth}{!}{%
\begin{tabular}{l *{5}{c}}
    \toprule
    \multicolumn{1}{c}{\textbf{Settings}}
    & \multicolumn{1}{c}{\textbf{LLaVA-1.5-7b}}
    & \multicolumn{1}{c}{\textbf{LLaVA-1.5-13b}}
    & \multicolumn{1}{c}{\textbf{LLaVA-NeXT-7b}}
    & \multicolumn{1}{c}{\textbf{LLaVA-NeXT-13b}}
    & \multicolumn{1}{c}{\textbf{XComposer-2.5-7b}}\\
    \midrule
    MT-VQA-v2 & 70.65 & 72.94 & 75.18 & 75.26 & 75.76 \\
    MT-GQA $\rightarrow$ MT-VQA-v2 & 69.06 & 71.89 & 73.61 & 73.25 & 74.41 \\
    MT-GQA & 58.43 & 59.48 & 62.70 & 63.12 & 58.68 \\
    MT-VQA-v2 $\rightarrow$ MT-GQA & 57.45 & 58.78 & 61.43 & 62.60 & 58.06 \\
    \bottomrule
\end{tabular}%
}
\label{tab:transfer}
\end{table*}
\begin{table}[!t]
\centering
\setlength{\tabcolsep}{6pt}
\caption{Transfer results on MT-Video-MME. Average accuracy is reported across different methods.}
\resizebox{\linewidth}{!}{%
\begin{tabular}{l *{5}{c}}
    \toprule
    \textbf{Metrics} 
    & \textbf{Base} 
    & \textbf{Random} 
    & \textbf{Sample} 
    & \textbf{FastV} 
    & \textbf{Ours} \\
    \midrule
    $Acc_1$ & 44.3 & 26.8 & 25.5 & 26.2 & 28.5 \\
    $Acc_2$ & 46.7 & 25.3 & 26.4 & 27.3 & 27.3 \\
    $Acc_3$ & 48.2 & 30.7 & 31.3 & 31.6 & 34.6 \\
    $Avg$   & 46.4 & 27.6 & 27.7 & 28.4 & 30.1 \\
    \bottomrule
\end{tabular}
}
\label{tab:video}
\end{table}
\Cref{tab:transfer} reports more transfer validation experiments across LVLMs on MT-VQA-v2 and MT-GQA, showing that \method does not heavily depend on the specific training dataset, thereby demonstrating robust generalization ability.
Furthermore, \Cref{tab:video} reports transfer results on video question answering task. In detail, we transformed the video QA dataset Video-MME~\cite{fu2025video} into a 3-turn dialogue version, referred to as MT-Video-MME (including 500 dialogs for validation), and conducted comparative evaluations against baseline methods at a 70\% compression rate. Since XComposer-2.5-7B natively supports video input, we directly use the pre-trained weights obtained from training on the small combined dataset of MT-VQA-v2 and MT-GQA (only around 20k samples in total as we mentioned) to evaluate on the MT-Video-MME benchmark. As shown in the table, \method outperforms baseline approaches even without any task-specific training on MT-Video-MME, further demonstrating its strong transferability.

\section{More Visualizations}
\begin{figure*}[!p]
\centering%
\rotatebox{270}{\includegraphics[width=0.9\textheight]{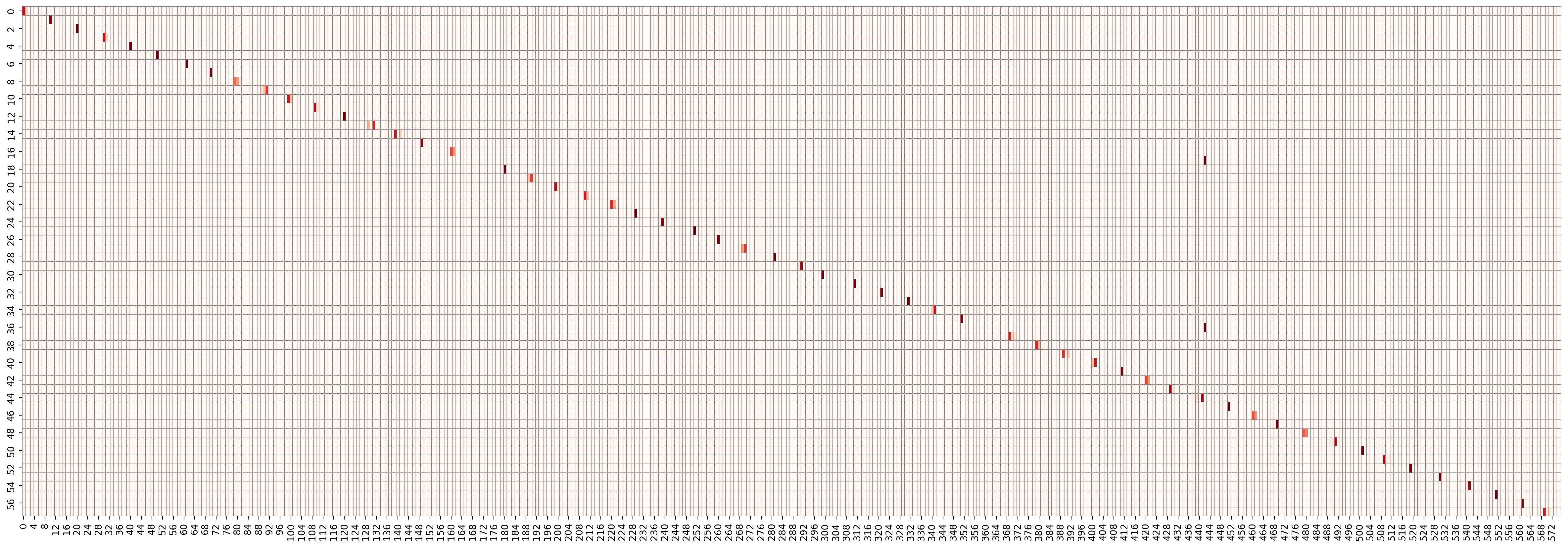}}%
\hspace{1cm}%
\rotatebox{270}{\includegraphics[width=0.9\textheight]{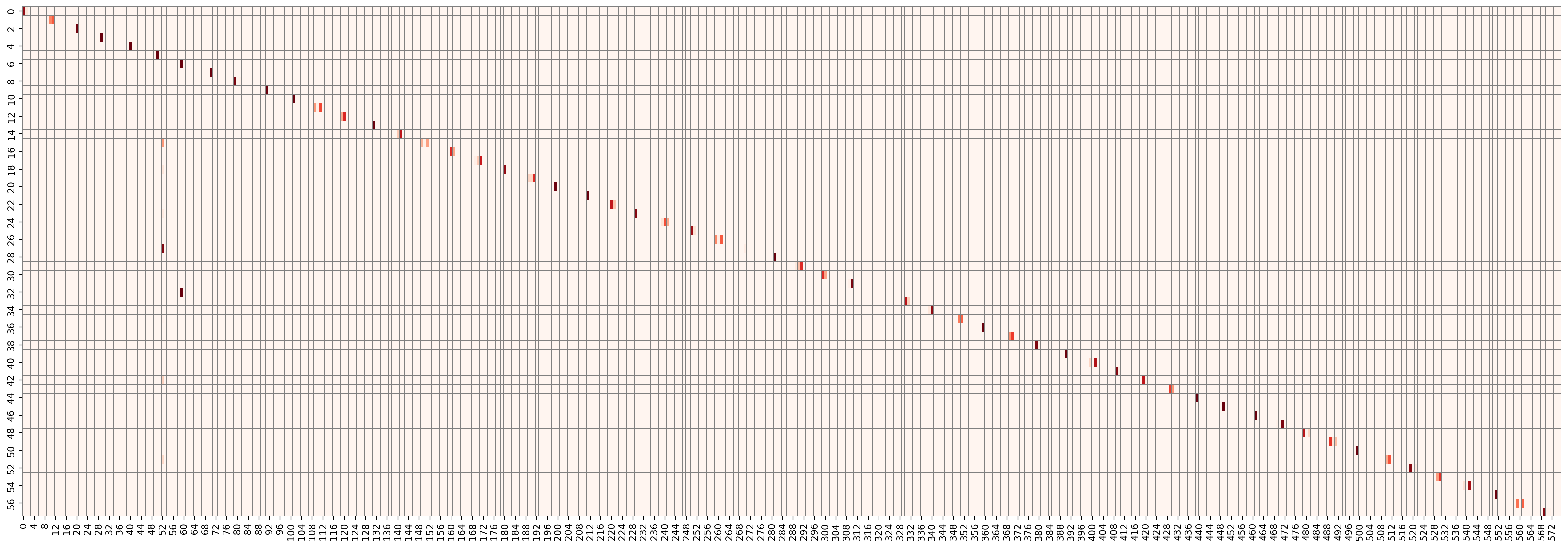}}%
\hfill%
\caption{Visualization of the compression projection for LLaVA-1.5-7b on MT-GQA with the compression rate of 90\%.}%
\label{fig:vis-llava-1.5-7b}%
\end{figure*}

\begin{figure*}[!p]
\centering%
\rotatebox{270}{\includegraphics[width=0.9\textheight]{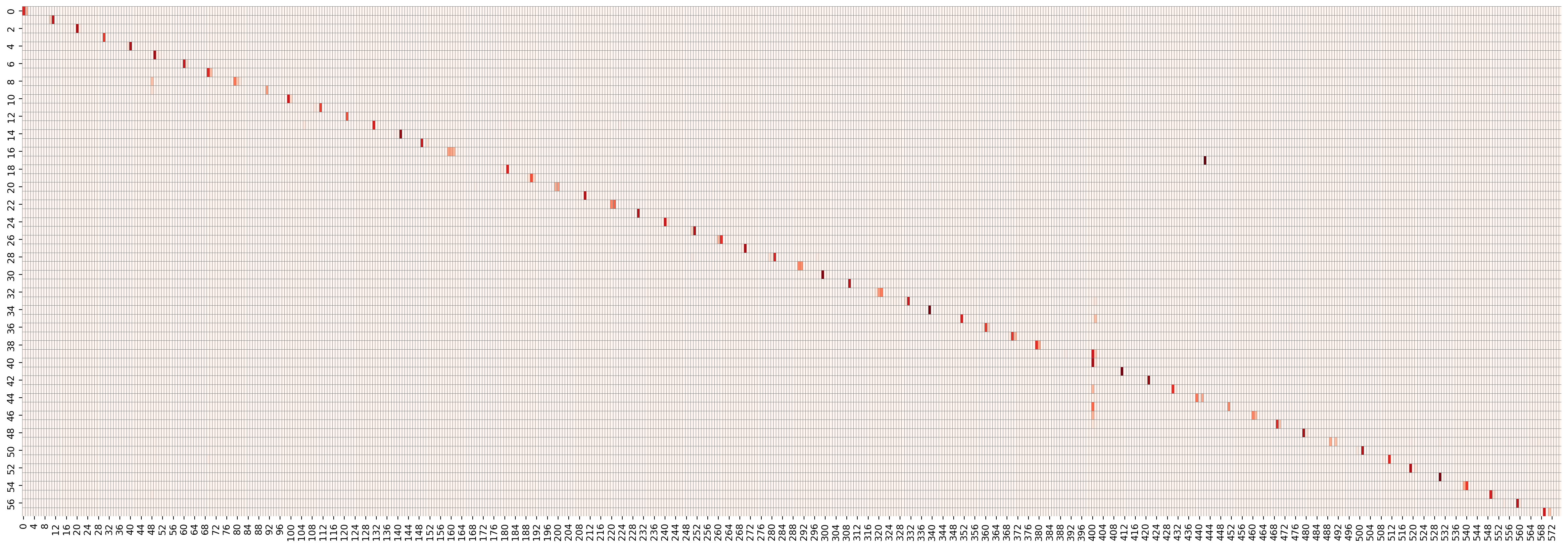}}%
\hspace{1cm}%
\rotatebox{270}{\includegraphics[width=0.9\textheight]{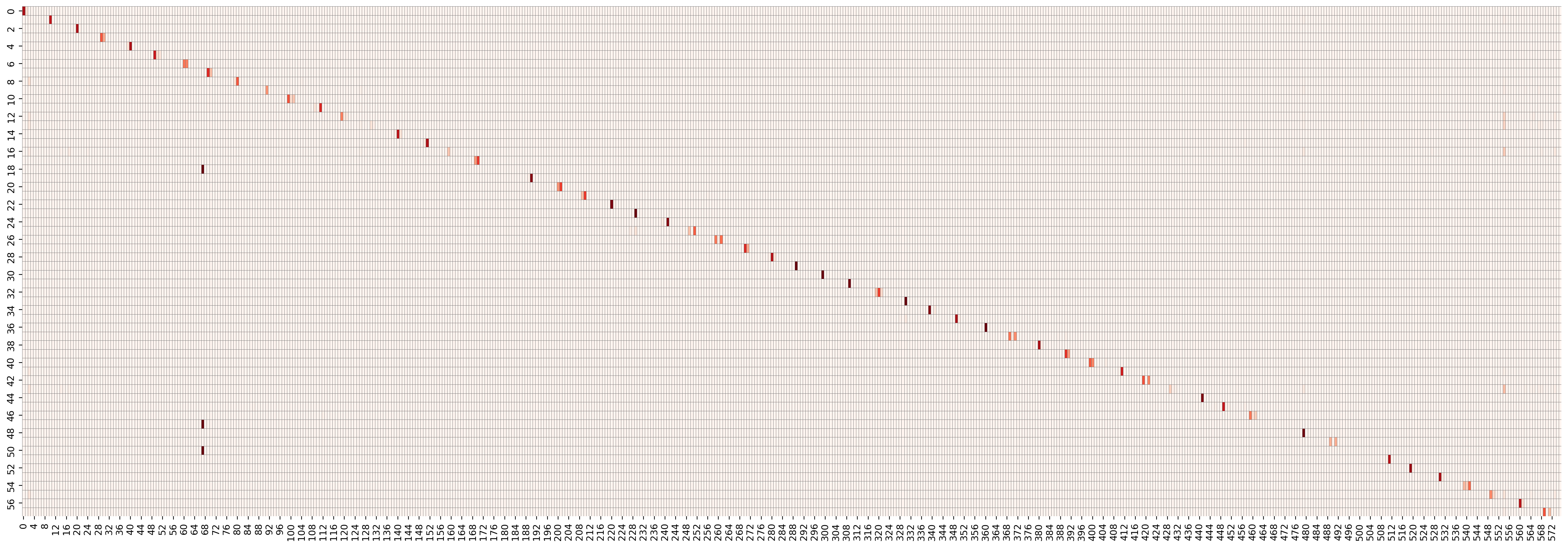}}%
\hfill%
\caption{Visualization of the compression projection for LLaVA-1.5-13b on MT-GQA with the compression rate of 90\%.}%
\label{fig:vis-llava-1.5-13b}%
\end{figure*}

\Cref{fig:vis-llava-1.5-7b,fig:vis-llava-1.5-13b} show the visualization of the generated compression projection for LLaVA-1.5-7b and LLaVA-1.5-13b on MT-GQA with a compression rate of 90\% (we randomly select two images as examples).
The row and column indices in the figures represent the original and reduced token indices, respectively, with darker colors indicating higher retention weights.
As observed, \method performs pruning and merging operations at different positions, but is primarily based on equidistant down-sampling, with specific adaptations for certain tokens.

\section{Discussions} \label{sec:discussions}
Most data-driven approaches for efficient model inference primarily focus on model pruning~\cite{ma2023llmpruner,sun2023simple}, efficient attention mechanisms~\cite{shen2021efficient,liu2024linfusion}, and efficient model architectures~\cite{liu2025nvila,vasu2025fastvlm,li2025tokenpacker}, particularly in designing vision encoders for stronger and more compact visual representations. However, these methods typically require fine-tuning the entire model, resulting in substantial computational overhead. In contrast, our proposed \method trains only a small number of lightweight linear projection layers ($D_q$, $D_k$, and $w$), yet it surpasses existing token reduction approaches. Owing to its efficiency, \method requires only a modest amount of training data (approximately 20k samples) while exhibiting strong transferability across datasets, as demonstrated in our transfer experiments. This generalization capability stems from the fact that \method is trained to preserve as much general visual information as possible for multi-turn dialogues rather than being specialized for specific image domains.

\end{document}